\documentclass{article}

\usepackage[preprint,nonatbib]{neurips_2019}

\usepackage[utf8]{inputenc} 
\usepackage[T1]{fontenc}    
\usepackage{hyperref}       
\usepackage{url}            
\usepackage{booktabs}       
\usepackage{amsfonts}       
\usepackage{nicefrac}       
\usepackage{microtype}      
\usepackage[backend=bibtex]{biblatex}
\bibliography{neurips_2019}

\usepackage{graphicx}
\usepackage{caption}
\usepackage{subcaption}
\usepackage{qtree}
\usepackage{mathtools}

\title{On the Possibility of Rewarding\\ Structure Learning Agents:\\Mutual Information on Linguistic Random Sets}

\author{%
  Ignacio Arroyo-Fern\'andez\thanks{Work developed as a member of the I+D+I laboratory at GACS asesores financieros (CDMX headquarters).} 
  \\
  Postgraduate Studies Division\\
  Universidad Tecnol\'ogica de la Mixteca\\
  Oaxaca, Mexico 69000 \\
  \texttt{iaf@mixteco.utm.mx} \\
   \And
  Mauricio Carrasco-Ruíz\thanks{We thank the \textit{``Laboratorio universitario de cómputo de alto rendimiento''} (LUCAR--IIMAS--UNAM) for providing computing nodes.} \\
  Facultad de Ciencias -- UNAM \\
  Ciudad Universitaria, CDMX \\
  \texttt{maucarrui@ciencias.unam.mx} \\
  \AND
  J. Anibal Arias-Aguilar \\
  Postgraduate Studies Division\\
  Universidad Tecnol\'ogica de la Mixteca\\
  Oaxaca, Mexico 69000 \\
  \texttt{anibal@mixteco.utm.mx} \\
}

\begin{document}

\maketitle

\begin{abstract}
We present a first attempt to elucidate a theoretical and empirical approach to design the reward provided by a natural language environment to some structure learning agent. To this end, we revisit the Information Theory of unsupervised induction of phrase-structure grammars to characterize the behavior of simulated actions modeled as set-valued random variables (random sets of linguistic samples) constituting semantic structures. Our results showed empirical evidence of that simulated semantic structures (Open Information Extraction triplets) can be distinguished from randomly constructed ones by observing the Mutual Information among their constituents. This suggests the possibility of rewarding structure learning agents without using pretrained structural analyzers (oracle actors/experts).
\end{abstract}

\section{Introduction}
An agent structuring its perceptions is a key problem in Artificial Intelligence. Significant advances have been observed in this matter recently and a number of applications have been potentiated with the rising of Deep Neural Network models (e.g. semantic segmentation \cite{long2015fully,wei1997automatic}). These models have demonstrated their ability to abstract the perceptions of autonomous agents who perform well-defined (or implicitly defined) tasks in virtual and real environments \cite{ros2015vision,eysenbach2018diversity}. The case of environments in which humans can abstract explicit knowledge that can be easily transferred to agents in the form of goals has been well studied, although there are multiple open problems such as the partial observability of states, variance in learning and generalization to unknown environments \cite{duan2016rl,sutton2018reinforcement}. In addition, model interpretability and environment structure become central issues. This is because intelligence tests generally consider agents that already acquired knowledge, but how an agent can be pre-configured to be able to acquire knowledge about its environment (making it \textit{potentially intelligent}) remains a barely explored question \cite{bengio2019meta,gershman2010learning}.

Special cases of the above questions arise when exploring natural language environments, where the objectives of the tasks cannot be explicitly defined. Up to our knowledge, such cases are not widely studied until now. Thus, very interesting tasks such as dialogue generation remain as open problems \cite{li2016deep,serban2017multiresolution} (although the availability of training data is not a problem, e.g. online chats). This is mainly due to the partial (or nonexistent) observability of the statistical structure of language. This limits the kind of studied tasks to those where distant supervision is available, as occurs in text-based games (i.e. the task provides pairs of language instructions and goals) \cite{choi2017coarse,luketina2019survey}, or where pretrained structure analyzers (experts or static oracles) are available \cite{he2016deep,narasimhan-etal-2016-improving} (this makes agents highly language dependent), or where much less detailed information is asked to the agent (e.g. information retrieval \cite{hofmann2011contextual,narasimhan-etal-2016-improving}).
Recently, the authors of \cite{bouchacourt2019miss} studied messages and responses shared by cooperating agents given artificially constructed environments that require semantic understanding of tasks. Authors observed that the Mutual Information (MI, \cite{shannon1948mathematical}) between orders describing fruits and the traits of the agent's actions (selected fruits) correlates with the failure and success of goals.

In this paper, we study the behavior of MI computed from gold standard actions taken by a simulated Open Information Extraction (OpenIE \cite{banko2007open}) agent, and compared them with the actions of a random agent. The aim of this comparison is to observe how such agents differ in the sense of the \textit{unsupervised induction of semantic structures}, which is a concept we extrapolated from \cite{de1999unsupervised}. Apart from the problem of studying the possible rewards a natural language environment may provide, an important challenge for us has been that linguistic samples are not ordinal. Also sets of them induce high algorithmic complexity if they are treated as joint distributions of exact observations or points. Therefore, we needed to rethink probabilities in the sense of random sets \cite{cressie1987random,molchanov2006theory}, assuming that linguistic random sets meet axioms of Borel algebras and measures. Thus, our ongoing work tries to elucidate the possibility of using MI to reward set-valued agents' actions while learning semantic structure and thus to propose a rich variety of \textit{Semantic Reinforcement Learning} problems.

\section{Background}

\paragraph{Information Theory of Phrase-Structure.}
De Marcken \cite{carl1996marcken,de1999unsupervised} provided insight on how entropy differences in phrase structure can be used to induce grammar in natural languages \cite{ARROYOFERNANDEZ2019107,hale2003information}. A case that is particularly interesting for our purposes is that of rules producing prepositional phrases like $P\to V P N$, where $V$ is a verb phrase and $N$ is a noun phrase, both them linked by another prepositional phrase $P$. For such kind of rules, regularly it holds that $I(V, P)>I(P, N)>I(V, N)$, where $I(A, B)$ is the Shannon's Mutual Information between the distribution of $A$ and the distribution of $B$. The reasoning underlying such a hypothesis is that, in English, verbs and prepositions are semantically more associated than prepositions and nouns. 

Although De Marcken used a corpus of minimal phrases manually selected, we think his hypothesis holds for structures of any complexity (at the end, language structure is recursive \cite{doi:10.1080/09296171003643189,chomsky1959algebraic,levy2004neuro}). That is, predicates (verb phrases) filter their possible arguments (noun phrases), which defines the so called selectional preferences (or thematic relations) guided by semantics \cite{katz1963structure,paducheva-1991-semantic,resnik1993selection}. Therefore, in this paper we generalize such hypothesis to introduce an Information-Theoretic perspective of semantic structure induction from open domain and unlabeled text. For instance, the phrase ``The adventures of Alice in Wonderland'' can be structured semantically as follows:
\begin{equation}
    \label{eq:example1Tree}
    \begin{scriptsize}
    \Tree [.$T_t$ [.$Y$ $Y_t:\text{of}$ ] [.$X,Z$ [.$X$ $X_t:\text{The adventures}$ ] [.$Z$ $Z_t:\text{Alice\hspace{1mm}in\hspace{1mm}Wonderland.}$ ] ] ]
    \end{scriptsize}
\end{equation}
In Tree (\ref{eq:example1Tree}), $T_t$ models a semantic triplet, and we propose that its constituents $X,Y,Z$ are random sets taking values $X_t,Y_t,Z_t$, respectively. Here, e.g., $X$ is a linguistic random set and we measure the probability of it to contain the items $\{\omega_{1},\dots,\omega_{|X_t|}\}=X_t$. Notice that this latter is not a joint probability, i.e. $P(X_t)\neq P(\omega_{1},\dots,\omega_{|X_t|})$, but rather the probability of a set. $T_t$ can be modeled as the factors of a joint probability
\begin{equation}
    \label{eq:joint_prob}
    P(X,Y,Z)=P(Y|X,Z)P(Z|X)P(X),
\end{equation}
\noindent which would be the objective of some structure learning agent that also must learn the order of the constituents. Our extrapolation of the idea De Marcken proposed for grammar induction, is that semantic structures like $T_t$ can be characterized by the inequality (\ref{eq:iformation_ineq}):
\begin{equation}
    \label{eq:iformation_ineq}
    I(X,Y) > I(Y,Z) > I(X,Z).
\end{equation}
It says that when (\ref{eq:joint_prob}) has maximum likelihood for sets $X=X_t, Y=Y_t, Z=Z_t$ ($\forall T_t\in A_k$ and $t=1,\dots,|A_k|$), the information that can be gained about the dependence between $X$ and $Z$ is less than the information that can be gained about the dependence between $Y$ and $Z$, which is in turn less than the information that can be gained about the dependence between $X$ and $Y$. In this framework, $T_t$ is a subject-predicate-object triplet, where $X$ is the subject phrase, $Y$ is the predicate phrase and $Z$ is the object phrase. We explore this fact in characterizing structures of a gold standard dataset of semantic triples built by a hypothetically well-trained agent (an OpenIE algorithm). Notice that (\ref{eq:iformation_ineq}) provides an unsupervised constraint to obtain a maximum likelihood estimate of (\ref{eq:joint_prob}). This therefore suggest to investigate a natural language environment rewarding structure learning agents without the need for pretrained structure analyzers.

\paragraph{Probability on Linguistic Random Sets.}
Let $W$ be a random variable (a linguistic random set) taking values $W_1, W_2,\dots$ We assume that each measurable set $W_i$ of countable elements $\omega\in\Omega$ is in a $\sigma-$algebra $\Lambda$. To generate linguistic random sets $W=W_i$, there are two underlying mathematical mechanisms. The first one is to build a basis $\mathcal{V}=\{\Lambda_i,\Lambda^i\}$, where $\Lambda_i = \{W_j\in\Lambda | W_j\cap W_i\neq\emptyset\}$ and $\Lambda^i = \{W_j\in\Lambda | W_j\cap W_i=\emptyset\}$ for all $W_i$ we provide \cite{christian2002set}. Thus, we have a hit-and-miss topological space in which any $W_i'$ generates via finite unions $W_i'=\cup_{W_j\in\mathcal{V}_i} W_j$, where $\mathcal{V}_i\subset\mathcal{V}$. The second mechanism, requires to define a metric. The symmetric difference metric $h(W_i, W_{j})=|W_i\triangle W_{j}|=|W_i\cup W_{j}\setminus W_i\cap W_{j}|$, also called Fréchet--Nikodym or Hamming metric \cite[page 53]{bogachev2007measure}, is well-suited because it naturally works as the area of an indicator function $\mathbf{1}_{W_i\triangle W_j}$. This way, we can measure set lengths, and we have now a measure space $(\Omega, \Lambda, h)$. The elements of this space are the possible values of $W$, only those for which $h$ is a valid metric \cite{yon2016approximating}. Herein, any $W_i'$ is generated by the metric-induced topology. Next, according to 
\cite{bogachev2007measure,cressie1987random,molchanov2006theory}, 
we can define a \textit{capacity functional} in terms of $h$:
\begin{equation}
    \label{eq:capacity}
    P(W=W_i)=\int_\Lambda f(W_i,W_j)h(W_j)=\mathbb{E}_{W_j\in\Lambda}[f(W_i,W_j)],
\end{equation} 
which induces the probability space $(\Omega,\Lambda, P)$, where $h(W_j)=\int \mathbf{1}_{W_j}dh$ is the length of $W_j$. 
Eq. (\ref{eq:capacity}) can easily be generalized to the case of joint and conditional distributions. At the phrase level, elements $\omega,\omega'$ such that $P(\omega\in W_i|\omega'\in W_i)\to P(\omega\in W_i ,\omega'\in W_i)/P(\omega'\in W_i)$ are capable (probable) of building elements of semantic structures. The opposite occurs for elements $\omega,\omega'$ such that $P(\omega\in W_i|\omega'\in W_i) \to P(\omega\in W_{i})$. At the semantic structure level, dependencies of the form $P(W_i|W_j)$ follow from a similar reasoning for the sets involved. Notice that Eq. (\ref{eq:capacity}) has nice mathematical properties, e.g. it can be seen as a projector of the $W_j$s onto $W_i$, where $W_i$ is a set-valued filter. Therefore, to compute some $P_\varsigma(X=X_t)$ we build a Gaussian kernel estimator defined on set-valued agent's actions:
\begin{equation}
    \label{eq:guassian_hamming}
    P_\varsigma(X=X_t) = \mathbb{E}_{X_{t'}\in A_k}[ f_\varsigma(X_t,X_{t'})]= \mathbb{E}_{X_{t'}\in A_k}\left[\exp\left(\frac{-h(X_t,X_{t'})^2}{2\varsigma^2}\right)\frac{1}{\sqrt{2\pi\varsigma^2}}\right],
\end{equation}
\noindent where $\varsigma$ is the bandwidth \cite{rosenblatt1956remarks}.
By the kernel and expectation properties, (\ref{eq:guassian_hamming}) is a weighted sum and then an embedding function \cite{christian2002set,labuschagne2007vector}, yielding the probability space $(\Omega,\Lambda,P_\varsigma)$. Accordingly, the Shannon entropy defined on sets is: $H(X)=-\sum_{X_t\in A_k}P_\varsigma(X_t)\log P_\varsigma(X_t)$. Notice that $\varsigma$ can be used to control the \textit{strictness of observing} $X=X_t$ \cite{zadeh1968probability}, hence the behavior of $H(X)$. 

In the particular case of a simulated agent (either random or OpenIE) building a set of trees $A_k\ni T_t$, each action amounts to one tree $T_t$, and $|A_k|$ multiple actions can be taken at each step $k$. 
With this in mind, now we can evaluate (\ref{eq:iformation_ineq}) 
for the set-valued three (\ref{eq:example1Tree}) using the MI: $I(W,W')=H(W')-H(W'|W)\Longrightarrow$
    $H(Y)-H(Y|X) > H(Z)-H(Z|Y) > H(Z)-H(Z|X).$
 Similar ideas can be found in \cite{cakir2017mihash,perotti2015hierarchical,6984550}.
\section{Experiments and Results}
\label{sec:experiments_results}

\paragraph{Experimental setup.} Our experiments were conducted on the well known text dataset called Twenty News Groups dealing with different topics \cite{Lang95}. This is because we needed to observe that MI measurements hold regardless of the topics. To this end, we first processed a set of documents dealing with very similar topics, i.e. \verb+comp.os.ms-windows.misc+, \verb+comp.sys.ibm.pc.hardware+, \verb+comp.sys.mac.hardware+ and \verb+comp.windows.x+. After that, we processed a set of documents dealing with unrelated topics, i.e. \verb+talk.politics.guns+, \verb+sci.med+, \verb+sci.space+, \verb+rec.sport.hockey+, \verb+soc.religion.christian+, \verb+misc.forsale+, \verb+rec.autos+. Both these subsets were shuffled in order to simulate 
contexts (rather than states) drawn from text environments (i.e. agent's actions do not influence environment's state). In order to simulate a gold standard of semantic triples of an hypothetically well trained agent we used Standord CoreNLP\footnote{We used this system as a well-documented baseline; however, we prepared a much more extended version of our work involving different OpenIE systems and other hyperparameters.} configured to keep only the main triple from each clause observed \cite{manning-EtAl:2014:P14-5}. Also a non trained agent was simulated by extracting contexts of fixed length ($=10\sim$ average sentence length in words), where each context was split at two uniformly distributed indices with nonempty segments. The amount of step samples was fixed to $|A_k|=$100 and we simulated $k=1,\dots,$120 steps. The bandwidth to which a linguistic random set $W$ can be observed was set to $\varsigma=$5.0 and its elements are (1-3)-grams (of characters). Due to space constraints, these hyperparameters were selected by intuition and we directly used the unrelated topics dataset as development data to observe that the behavior of the variables holds similar to that observed in similar topics. 
\begin{figure}
\centering
\begin{subfigure}{.5\textwidth}
  \centering
  \includegraphics[trim={0 0 0 14mm},clip,width=.9\linewidth]{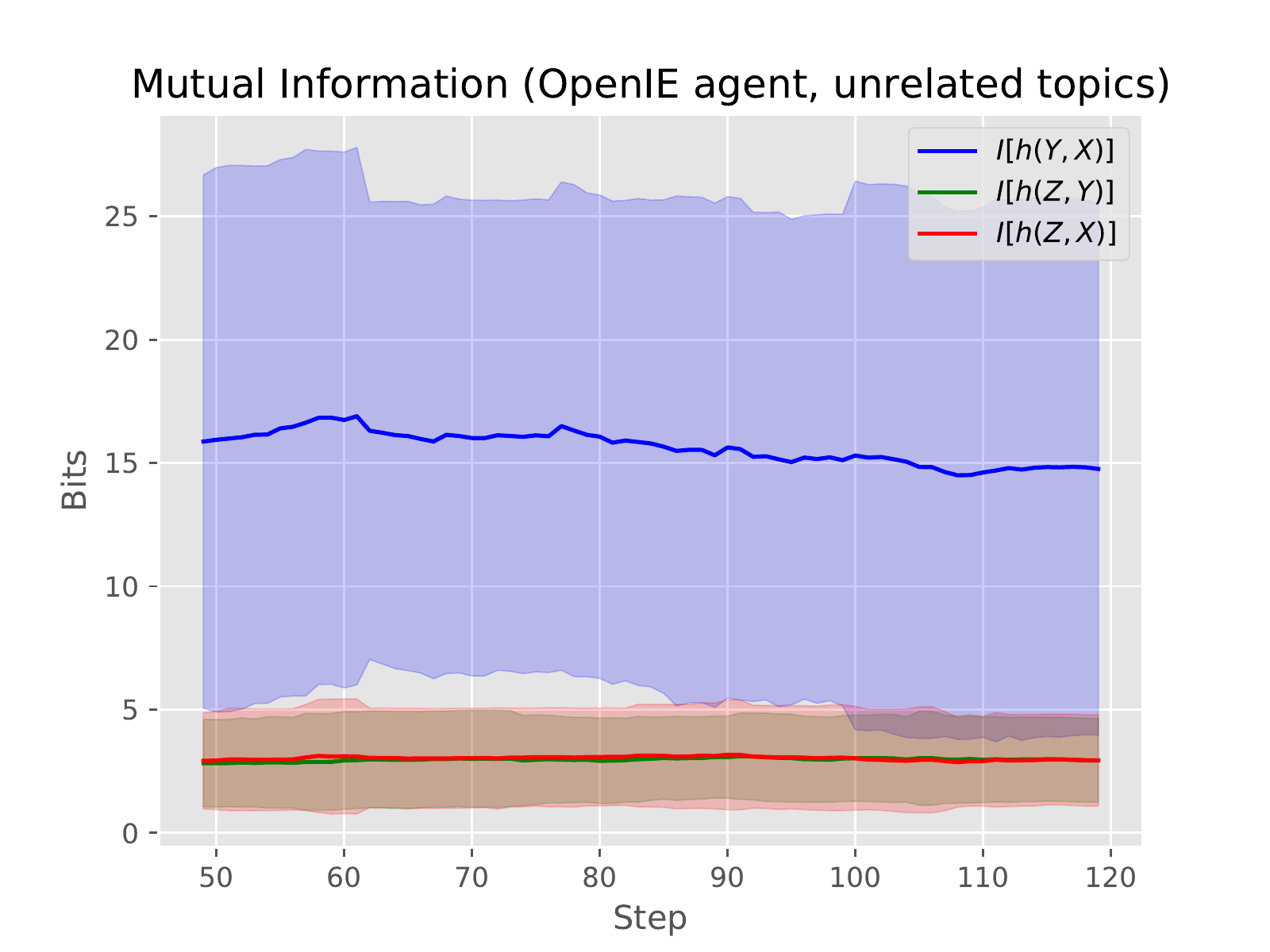}
  \caption{MIs on OpenIE set-valued agent actions.}
  \label{fig:dis_oie_mi}
\end{subfigure}%
\begin{subfigure}{.5\textwidth}
  \centering
  \includegraphics[trim={0 0 0 14mm},clip,width=.9\linewidth]{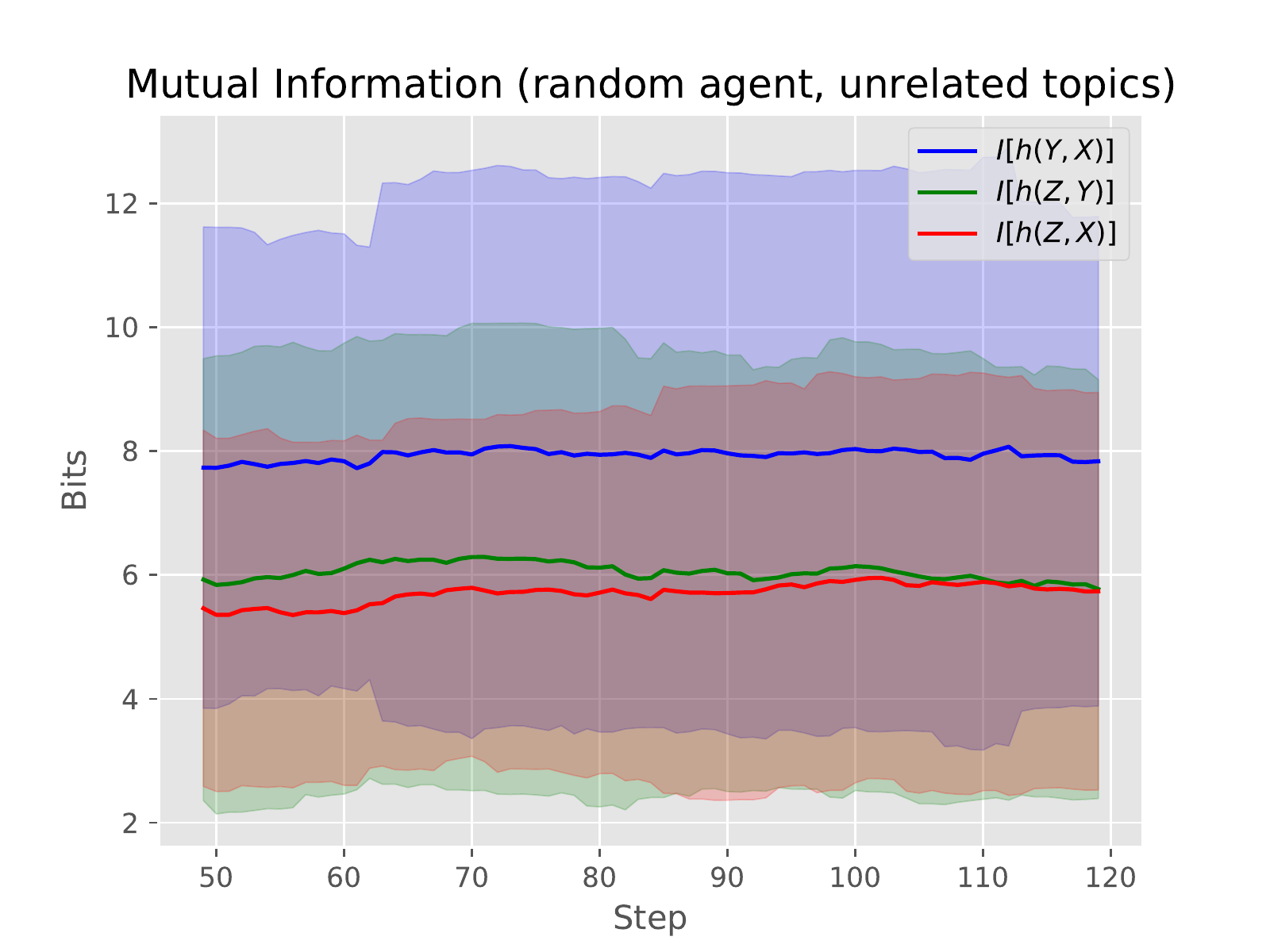}
   \caption{MIs for random set-valued agent actions.}
  \label{fig:dis_rdn_mi}
\end{subfigure}
\caption{MI for 120-step actions of simulated agents on unrelated topics (the same patterns were observed for similar topics), plotted using rolling mean of 50 steps.}
\label{fig:test}
\end{figure}
\paragraph{Results.} In Figure \ref{fig:dis_oie_mi} we observe the MIs $I(X,Y), I(Y,Z)$ and $I(X,Z)$ for our simulated gold standard agent (OpenIE) acting on randomized documents dealing with unrelated topics.\footnote{In the figures we denoted the Hamming metric inducing MI on the corresponding probability space by $I[h(W,W')]$.}\footnote{See \url{https://github.com/iarroyof/semanticrl/tree/master/figures} 
for similar topic plots.} It is remarkable that although there is a considerable variability of the MIs though the trajectory, the values of $I(X,Y)$ keep strongly separated (beyond its variance) from $I(Y, Z)$ and $I(X, Z)$. This indicates the correspondence between $Y$ and $X$ in the semantic structure, i.e. some well-trained structure learning agent can gain a meaningful amount of information from predicate phrases by observing subject phrases, while keeping predicate and subject phrases as independent as possible from object phrases. These findings are remarkable for us because we observed quite different behavior for the random agent in Figure \ref{fig:dis_rdn_mi}. In this case, the MIs were overall lower, as well as $I(Y, Z)$ and $I(X, Z)$ appeared to be free from the agent's influence, while approaching $I(X, Y)$ with high variability. 
\begin{figure}
\centering
\begin{subfigure}{.5\textwidth}
  \centering
  \includegraphics[trim={0 0 0 14mm},clip,width=.9\linewidth]{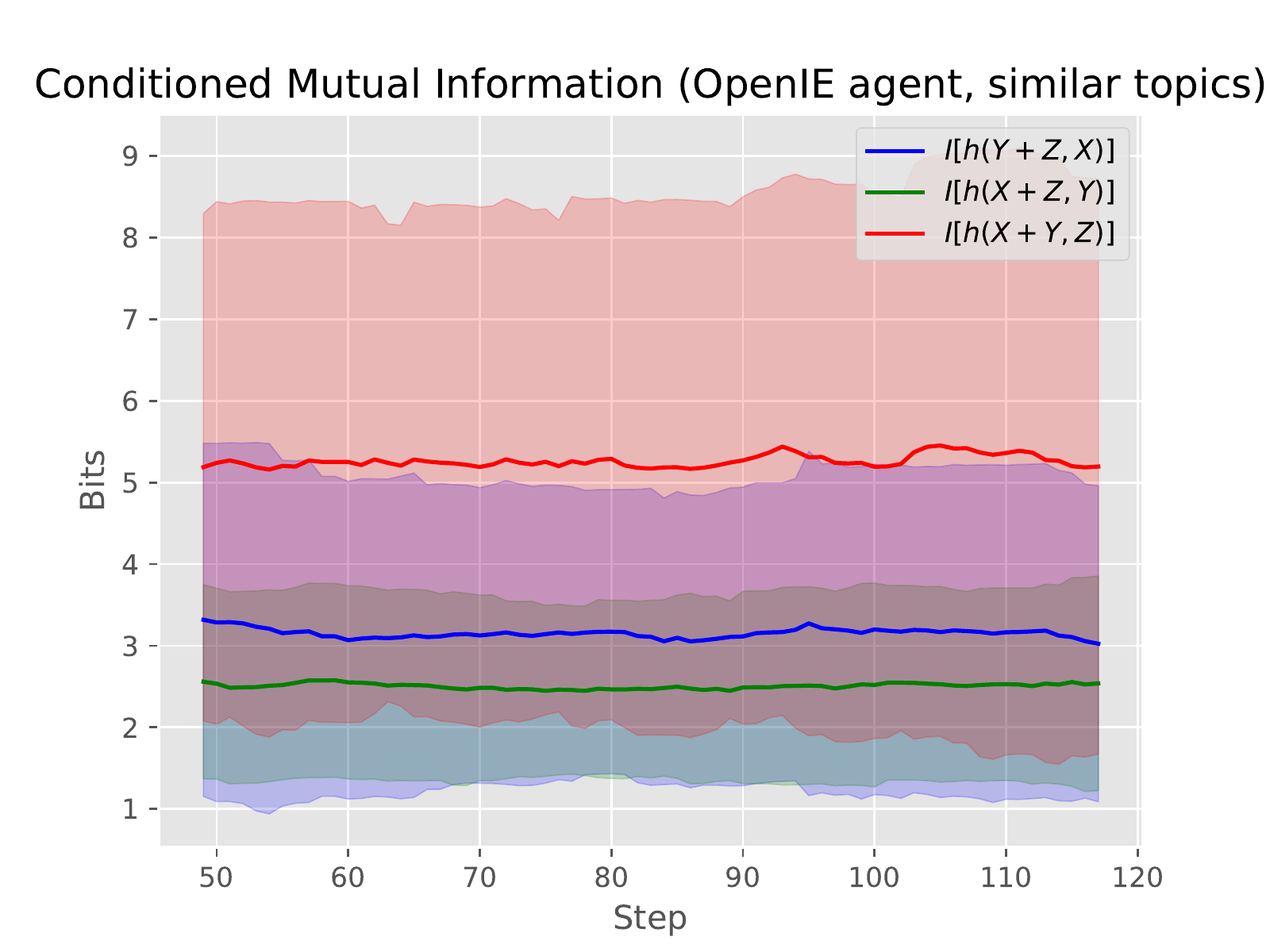}
  \caption{MIs for OpenIE set-valued agent actions.}
  \label{fig:sim_oie_cmi}
\end{subfigure}%
\begin{subfigure}{.5\textwidth}
  \centering
  \includegraphics[trim={0 0 0 14mm},clip,width=.9\linewidth]{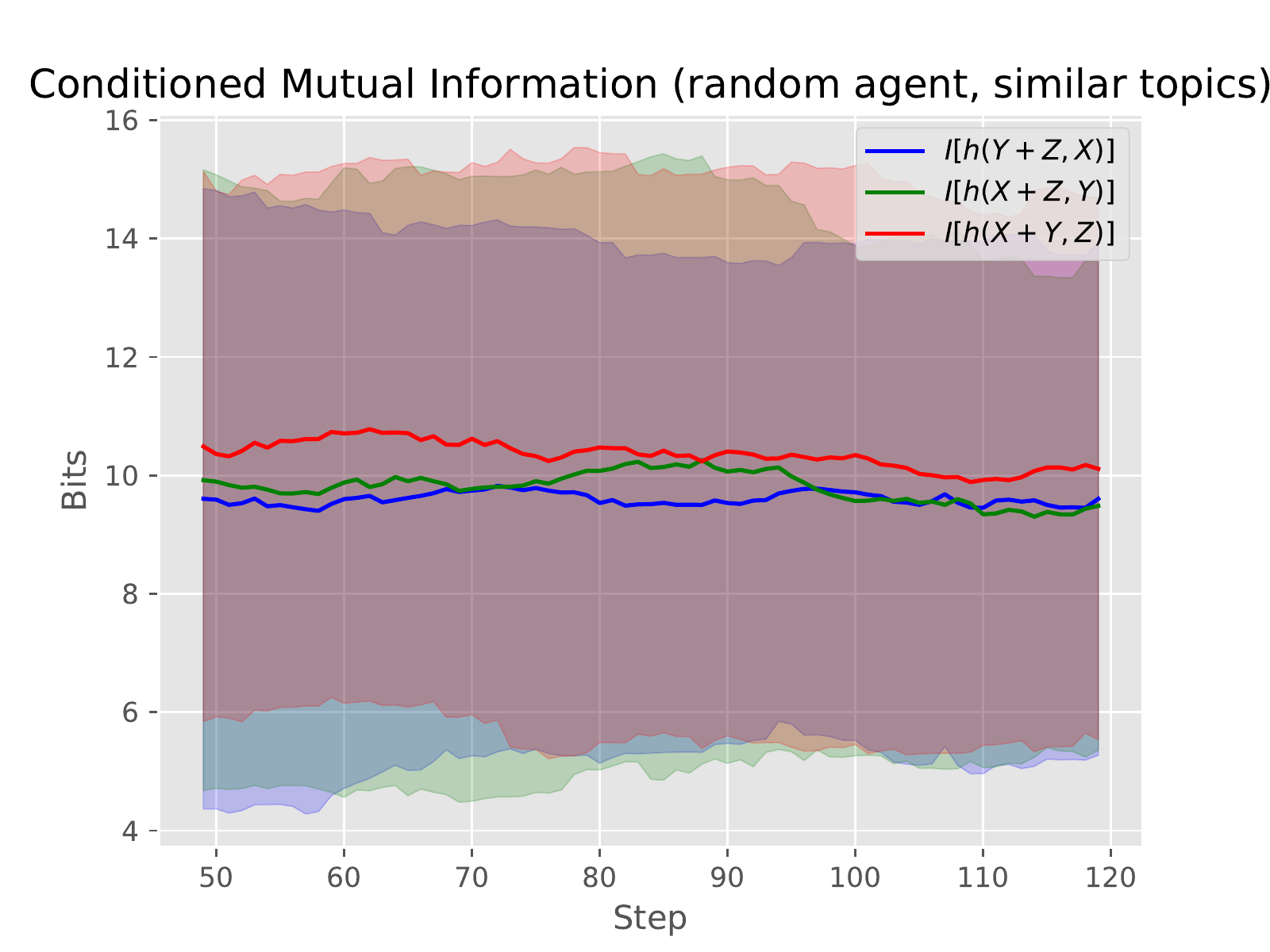}
   \caption{MIs for random set-valued agent actions.}
  \label{fig:sim_rdn_cmi}
\end{subfigure}
\caption{Joint-marginal MI for 120-step set-valued actions of simulated agents on similar topics, plotted using rolling mean of 50 steps.}
\label{fig:joints}
\end{figure}

We also explored the MI between joints and marginals, e.g. $I(X,Y;Z)$. This had the purpose of observing how the OpenIE agent gained information about the joint $X,Y$ from observing the marginal $Z$ (see Figure \ref{fig:joints}).\footnote{In the figures, we denoted the joint distribution by $W + W'$ given that corresponding set values were concatenated in our implementation, where it holds that $P(W, W')\leq P(W)\hspace{2mm}\forall W,W'\in(\Omega,\Lambda,h)$.} Although $I(X,Y;Z)$ couldn't be strongly distinguishable from the other MIs measured from the OpenIE agent (Figure \ref{fig:sim_oie_cmi}), we observed that the rolling means remained ordered. That is, $I(X,Y;Z)$ kept higher than the variance of the other two MIs. This is interesting because $I(X,Y;Z)$ means that the agent reaches the maximum information gained from the subject and the predicate phrases $X+Y$ by observing the object phrase $Z$. The opposite occurs if the agent tries to gain information from the subject and the object phrases $X+Z$, by observing the predicate phrase $Y$, i.e. $I(X,Z;Y)$. In the case of the random agent, no pattern was observed, see Figure \ref{fig:sim_rdn_cmi}. This agent seemed to be off, or simply inactive.

\section{Discussion}
Unlike to what we expected, $I(Z, Y)$ and $I(Z, X)$ couldn't be distinguished. That is, the uncertainty of $Z$ wasn't reduced semantically by observing either $Y$ or $X$ \cite{hale2006uncertainty}. This suggests that our extrapolation of De Marcken's ideas from the syntactic structure to the semantic structure needs to be explored in more detail, probably from a Causal Inference point of view. 

Despite the uncertainty shown by $Z$, we found it interesting that, in terms of the joint-marginal MIs, the maximum amount of information gained by the OpenIE agent was from the subject+predicate phrases $X + Y$, given that the object phrase $Z$ was observed (Figure \ref{fig:sim_oie_cmi}). We think $I(X,Y;Z)$ shows not only a quantity derived from language structure, but also it shows empirical evidence of selectional restrictions predicates impose to their possible arguments. It limits the semantic content of the resulting phrase, hence the entropy/uncertainty in the space of possible phrases is also limited. 

Another interesting point to discuss is the generality of 
the constraint (\ref{eq:iformation_ineq}), which clearly seems to limit the kind of structures an agent can learn. At the current stage of our work, we used this constraint as we already knew the OpenIE agent produces exclusively active voice structures. This motivates a reasonable discussion since an evident counterexample is passive voice, where the order of the structure elements change by focusing phrase meaning on the predicate. However, in the case of an structure learning agent being trained, we may define its action space in a convenient way such that the elements of any semantic structure are forced to fit the constraint. Here, we are assuming that the agent would end up acquiring some kind of transformational generative abilities.

\section{Conclusion}
Based on our initial experiments we observed that Mutual Information (in the sense of Shannon) provided a reliable framework to investigate the possibility of training structure learning agents whose actions are set-valued (linguistic random sets in the particular case of this paper). Although the De Marcken inequality did not hold completely, it is encouraging that our experiments showed that the MIs follow patterns that can be used to experiment with a number of reward functions (or even set-valued rewards). Nevertheless, much work is pending, including experiments with actual agents (e.g. \cite{arroyo2018unam}), different kernel estimators for action probabilities, different environment hyperparameters, and multiple Information-Theoretic metrics, as well as processing different data domains \cite{teboul2011shape}. 


\printbibliography

\end{document}